%% file: PAKDD23_GTEA.tex
\documentclass[runningheads]{llncs}
\usepackage[T1]{fontenc}
\usepackage{cite}
\usepackage{amsmath,amssymb,amsfonts}
\usepackage{algorithmic}
\usepackage{graphicx}
\usepackage{textcomp}
\usepackage{xcolor}
\def\BibTeX{{\rm B\kern-.05em{\sc i\kern-.025em b}\kern-.08em
    T\kern-.1667em\lower.7ex\hbox{E}\kern-.125emX}}
\usepackage{algorithm}
\usepackage{color}
\usepackage{ragged2e}
\usepackage{tabularx}
\usepackage{bm}
\usepackage{subfigure}
\usepackage{pgfplots}
\usepackage{xspace}
\usepackage{threeparttable}
\usepackage{amsmath}
\usepackage{epsfig}
\usepackage{booktabs}
\usepackage{multirow} 
\usepackage{arydshln}

\newcommand{\ttov}{\text{\textbf{T2V}}}

\newcommand{\Graph}{\ensuremath{\mathcal{G}}}
\newcommand{\V}{\ensuremath{\mathcal{V}}}
\newcommand{\E}{\ensuremath{\mathcal{E}}}
\newcommand{\vctr}[1]{\ensuremath{\mathbf{#1}}}

\newcommand{\mlp}{\text{MLP}}

\begin{document}
\title{GTEA: Inductive Representation Learning on Temporal Interaction Graphs via Temporal Edge Aggregation}
\titlerunning{GTEA: Inductive Learning on TIGs via Temporal Edge Aggregation}
%
\author{Siyue Xie \inst{1}$^\star$, Yiming Li \inst{1}\thanks{Equal contributions.}, Da Sun Handason Tam \inst{1} \and
Xiaxin Liu \inst{2}, Qiufang Ying \inst{2} \and
Wing Cheong Lau \inst{1}, Dah Ming Chiu \inst{1} \and
Shouzhi Chen \inst{2}
}

%
\authorrunning{S Xie, Y Li, et al.}
%

\institute{
  The Chinese University of Hong Kong \and Tencent Technology Co.Ltd
}

\maketitle              
\begin{abstract}
  In this paper, we propose the Graph Temporal Edge Aggregation (GTEA) framework for inductive learning on Temporal Interaction Graphs (TIGs).
  Different from previous works, GTEA models the temporal dynamics of interaction sequences in the continuous-time space and simultaneously takes advantage of both rich node and edge/ interaction attributes in the graph. 
  Concretely, we integrate a sequence model with a time encoder to learn pairwise interactional dynamics between two adjacent nodes.
  This helps capture complex temporal interactional patterns of a node pair along the history, which generates edge embeddings that can be fed into a GNN backbone.
  By aggregating features of neighboring nodes and the corresponding edge embeddings, GTEA jointly learns both topological and temporal dependencies of a TIG.
  In addition, a sparsity-inducing self-attention scheme is incorporated for neighbor aggregation, which highlights more important neighbors and suppresses trivial noises for GTEA.
  By jointly optimizing the sequence model and the GNN backbone, GTEA learns more comprehensive node representations capturing both temporal and graph structural characteristics.
  Extensive experiments on five large-scale real-world datasets demonstrate the superiority of GTEA over other inductive models.

\end{abstract}
\section{Introduction}

  Representation learning on temporal graphs is a hot topic in the community of graph learning, where researchers have devoted to mining temporal correlations from graphs and achieve great successes across different domains \cite{li2018dcrnn_traffic, zuo2018embedding, huang2019finding, qiu2020temporal}.
  However, many methods \cite{nguyen2018continuous,zhang2020learning} only work for a fixed topology (transductive settings), while in product scenarios, a temporal graph usually evolves as new nodes/ edges added, which requires a model to work inductively.
  Therefore, in this work, we focus on the inductive learning on Temporal Interaction Graphs (TIGs), where each edge includes all interaction records between two nodes over the history.
  Applications on TIGs are common in real-world environments, such as recommendation systems \cite{ma2020streaming}, social network analytics \cite{kumar2019predicting, rossi2020temporal}, etc.

\begin{figure}[t]
    \centering
    \includegraphics[scale=0.2]{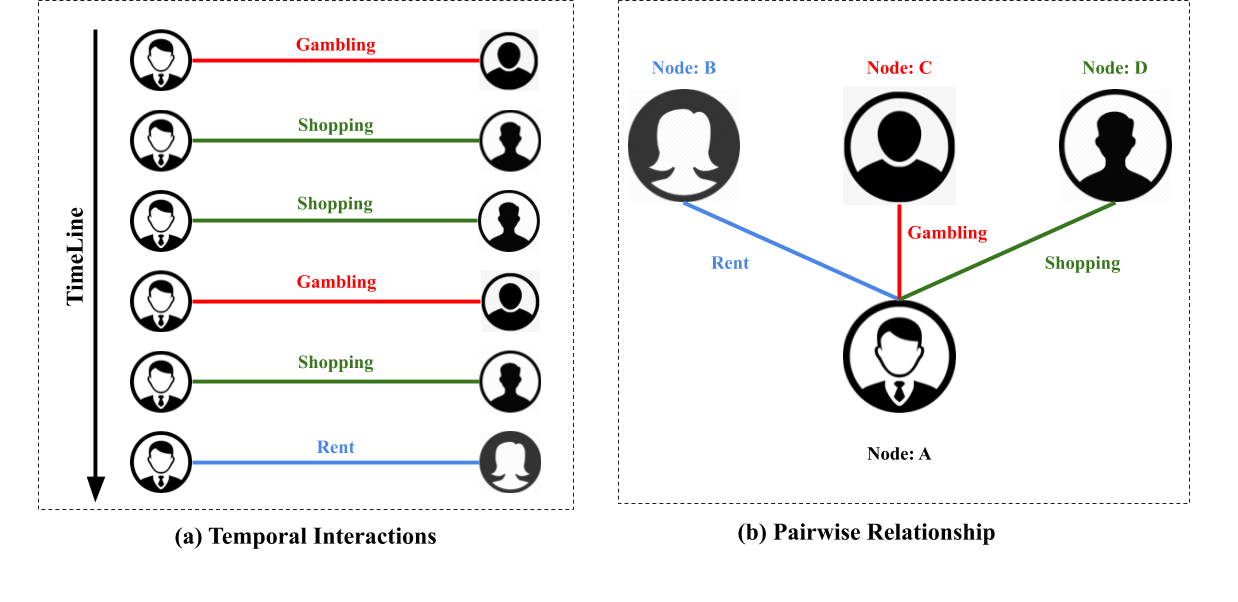}
    \caption{
      Motivations of capturing pairwise relationships: there are different kind of interaction between nodes in TIG, e.g., node $A$ behaves normally with node $B$ and $D$, while conducting (illicit) gambling activities (what we are interested in) with node $C$. 
    }
    \label{fig:example}
  \end{figure}

  Although researchers have made substantial advances in processing temporal graphs, it is still challenging to learn discriminative and fine-grained embeddings for TIGs.
  Previous works commonly preprocess a temporal graph by compressing time-related records within a regular time interval, which yields a spatial-temporal graph with multiple snapshots.
  However, interactions/ events in TIGs usually occur irregularly along time.
  Such snapshots are coarse approximations of temporal interactions and resulting in great losses of fine-grained temporal patterns, which prevents spatial-temporal methods \cite{yu2018spatio, singer2019node, li2018dcrnn_traffic} from being generalized to TIGs.
  Some works \cite{nguyen2018continuous, xu2020inductive} circumvent such a drawback by grouping all interactions associated with a node to form a consecutive time series for temporal dynamics analyses. 
  Although it preserves the time granularity of interactions, it mixes the temporal behaviors of different neighbors of a node and sometimes obfuscates some explicit temporal relationships between two nodes.
  Instead, modeling the interaction dynamics between a pair of adjacent nodes can be more helpful to capture temporal relation patterns from TIGs.
  As the example in Fig. \ref{fig:example}, a gambler involves in many interactions, but abnormal behaviors can be readily captured by the pairwise interaction dynamics between node $A$ and $C$, which in turn, helps identify the roles or illicit activities of these nodes. 

  Another drawback of previous works is that the edge information is usually underestimated or even ignored for graph learning.
  However, one should expect that edges carry rich interactional information of TIGs, which can be instrumental in the learning process.
  Following this intuition, some works \cite{gong2019exploiting,simonovsky2017dynamic} take edge information into account by concatenating both node and edge features for neighborhood aggregations.
  Others try to distinguish important connections by introducing dense attention mechanism \cite{velivckovic2017graph, zhang2018gaan, shi2020masked}.
  However, naive feature concatenations can be inferior for learning.
  Dense attention inevitably introduces noises during aggregation, which may overwhelm critical information in some tasks where only few neighbors are of interests (e.g., anomaly detection).

  To handle the aforementioned challenges, we propose \textbf{G}raph \textbf{T}emporal \textbf{E}dge \textbf{A}ggregation (GTEA) for inductive representation learning on TIGs based on Graph Neural Networks (GNN).
  Different from previous works, we present a new perspective to deal with TIGs.
  Instead of partitioning a temporal graph into multiple snapshots or grouping all related interactions of a target node to form a time series, we propose to mine pairwise interaction patterns from edges.
  Specifically, we adapt a sequence model (e.g., LSTM \cite{hochreiter1997long} or Transformer \cite{vaswani2017attention}) to mine the temporal interaction dynamics between two adjacent nodes.
  This helps capture complex interactional patterns of a node pair over the history.
  In addition, we integrate a time-encoding scheme \cite{mehran2019time2vec} with the sequence model, enabling GTEA to learn continuous and irregular time patterns for interaction events.
  To jointly learn topological dependencies and temporal dynamics, we utilize a GNN to capture the relationships among nodes, where embeddings outputted by the sequence model are taken as edge features and incorporated into the neighborhood aggregation process.
  Furthermore, we adapt a sparsity-inducing attention mechanism to augment the aggregation, which refines neighborhood information by filtering out noises raised by unimportant neighbors.
  By training GTEA in an end-to-end manner, all modules can be jointly optimized, which yields discriminative node representations for downstream tasks.
  Extensive experiments are conducted on five real-world datasets across different tasks, where results demonstrate the superiority of GTEA over other inductive models.
  The contributions of our work are summarized as follows:
  \begin{itemize}
    \item We present a novel perspective for modeling TIGs, which helps capture fine-grained interaction patterns from node pairs.
    \item We propose a general framework, GTEA, for inductive learning on TIGs, which yields discriminative node representations for downstream tasks.
    \item We conduct extensive experiments on five large-scale datasets. Experimental results demonstrate the great effectiveness of GTEA.
  \end{itemize}

  \section{Related works}

  \subsection{Temporal Dynamics Modeling on Graph-structured Data}
  
  Temporal graphs are ubiquitous in real-world scenarios, which motivates researchers to extend the target from learning from static graphs to the temporal domain \cite{nguyen2018continuous,qu2020continuous,qiu2020temporal}. 
  A common way is to form multiple static graph snapshots to approximate the time-continuous temporal graph \cite{yu2018spatio,singer2019node,li2018dcrnn_traffic}.
  However, critical details, e.g., the time granularity of interactions in TIGs, may lose in the simplification process.
  Instead of learning from snapshots, \cite{trivedi2017know, zuo2018embedding, trivedi2018dyrep} learn node representations using temporal point process.
  \cite{kumar2019predicting,zhang2020learning,ma2020streaming,rossi2020temporal} progressively update node embeddings as a new event/ interaction occurs.
  TGAT \cite{xu2020inductive} adopts a time-encoder with self-attention to aggregate interactions of neighbors.
  However, many of above models only work for transductive tasks, which restricts their generalization ability.
  Node embeddings updated event-by-event may be biased by noises and lose focus on the information of interest, e.g., the rare gambling activities as shown in Figure \ref{fig:example}.
  Although TGAT adopts attention on different interactions, it lumps all and may fail to distinguish the interactions of the same instance, e.g., in Figure \ref{fig:example}, there are two different gambling interactions but both are with node $C$.
  Different from previous methods, GTEA attempts to inductively model the temporal dynamics by looking at the complete interaction history between each pair of adjacent nodes, which enables it to capture specific and fine-grained mutual interaction patterns.
  We make more discussions and comparisons in Appendix 1 \footnote[1]{Appendix can be found in: https://github.com/xslangley/GTEA}.
  
  \subsection{Representation Learning on Graphs with Edge Features}

  Edges are natural reflections of relationships among instances but the information they carried is usually underestimated.
  Therefore, some pioneers propose to mine from edges to enhance a model, e.g., ECConv \cite{simonovsky2017dynamic} attempts to generate a set of edge-specific weights for Graph Convolutional Networks (GCN), while EGNN \cite{gong2019exploiting} constructs a weighted graph for each edge features' dimension.
  EdgeConv \cite{wang2019dynamic} and AttE2Vec \cite{bielak2020attre2vec} learns edge features to describe relationships between adjacent nodes.
  Instead of learning from an edge directly, CensNet \cite{jiang2020co} converts a original graph into a line graph, where edges are mapped as nodes in the new graph.
  Motivated by previous works, GTEA is extended to learn from edges in the temporal domain by modeling the mutual interaction dynamics, which improves the model performance and node representation power.

  \section{Proposed Methods}

  \subsection{Problem Formulation}

  A \textbf{Temporal Interaction Graph} (TIG) is an attributed graph $\Graph = (\V, \E)$  where $\V$ is a set with $N$ nodes and $\E$ a set with $M$ edges.
  A node $u$ is associated with features $\vctr{x}_{u} \in \mathbb{R}^{D_N}$, while an edge between $u$ and $v$ corresponds to a sequence of interaction events, denoted as $\{ \vctr{e}_{uv}^{k}=(t^k_{uv}, \vctr{f}^k_{uv}); k=1, 2,..., S_{uv} \}$, where $t^k_{uv}$ is the time stamp of the $k$-th event, $\vctr{f}^k_{uv} \in \mathbb{R}^{D_E}$ the interaction features and $S_{uv}$ the length of the sequence.
  Given the interaction history, the goal of GTEA is to learn discriminative node representations, which can be used in downstream tasks such as node classifications or future link predictions. 

  \subsection{Overview of GTEA}
  
  \begin{figure*}[t]
    \centering
    \includegraphics[scale=0.4]{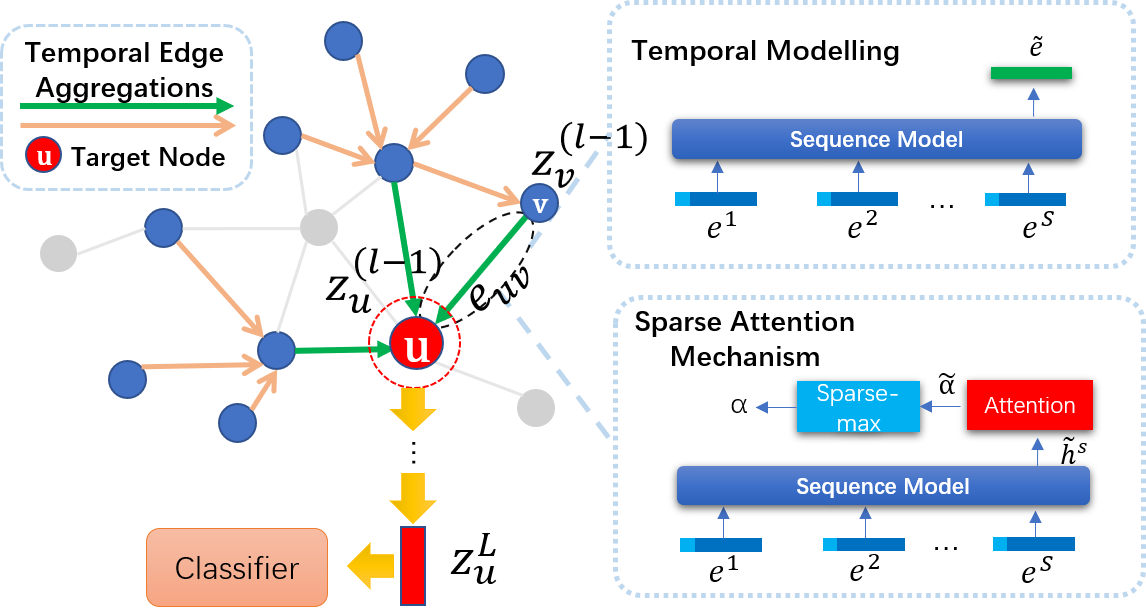}
    \caption{
      The framework of GTEA, where a sequence model enhanced by a time-encoder is proposed to learn embeddings for edges. The learned edge embeddings will be aggregated together with node attributes by the GNN backbone with a sparse attention mechanism, which helps yields discriminative node embeddings.
    }
    \label{fig_framework}
  \end{figure*}
  
  The architecture of GTEA is shown in Figure \ref{fig_framework}.  
  In TIGs, interaction events occur between two nodes from time to time, which motivates us to mine fine-grained interaction patterns from pairwise events. 
  Targeting on this goal, GTEA utilizes a sequence model to learn the dynamics of pairwise interactions to represent edges.
  An additional time-encoder is further introduced to capture irregular temporal patterns.
  Learned edge embeddings are fed into a GNN, along with a sparse-inducing attention mechanism for neighbor aggregations, which jointly captures both topological and time-related dependencies among nodes.
  With these designs, GTEA is able to yield discriminative representations for TIGs.

  \subsection{Learning Edge Embeddings for Interaction Sequences}

  \textbf{Interaction Dynamics Modeling with Sequence Models}~ 
  In TIGs, the types of interactions of a node involved can vary greatly with different neighbors.
  However, interaction patterns of two specific nodes are usually consistent even as time goes.
  Therefore, it is more reasonable to model the interaction behaviours edge-by-edge instead of mixing all interactions from different neighbors.
  Given the interaction history $[\vctr{e}_{uv}^1, ..., \vctr{e}_{uv}^{S_{uv}}]$ of edge $(u, v)$, we adopt a sequence model $Enc_{i}(\cdot)$ to learn the interaction dynamics as follows:
  \begin{align}
    \vctr{\tilde{e}}_{uv}=Enc_{i}([\vctr{e}_{uv}^1, ..., \vctr{e}_{uv}^{S_{uv}}]),
   \label{eq:seq-repr-ori}
  \end{align}
  where $Enc_{i}(\cdot)$ indicates the interaction encoder and $\vctr{\tilde{e}}_{uv}$ is the edge embedding to represent the interaction sequence.
  In our experiments, we implement $Enc_{i}(\cdot)$ by LSTM \cite{hochreiter1997long} and Transformer \cite{vaswani2017attention}.
  In LSTM, we represent $\vctr{\tilde{e}}_{uv}$ by the hidden output of the last time unit.
  As for Transformer, an interaction is correlated with all other interactions following self-attention.
  Therefore, it is sufficient for us to represent $\vctr{\tilde{e}}_{uv}$ by the embedding with respect to the last interaction.
  In this way, interactions of the same node pair can be completely reviewed by the sequence model, which helps capture specific interactional patterns for any two connected nodes.
  More technical details refer to Appendix 2.1.\\
  \\
  \textbf{Edge Feature Enhancement with Time Encoding}~ 
  Sequence models implicitly assume the time gap between consecutive inputs is regular along the timeline, while interactions happened in TIGs do not follow. 
  Therefore, to capture more complex time-related interaction patterns, we enhance GTEA by integrating $Enc_{i}(\cdot)$ with a time-encoder, which is adapted from Time2Vec ($\ttov$) \cite{mehran2019time2vec}.
  Specifically, for any given time $t$, a time embedding $\tau(t) \in \mathbb{R}^{l+1}$ can be generated through:
  \begin{equation}
    \label{eq:t2vec_def}
    \tau(t)[i]=
      \begin{cases}
        \omega_i t + \varphi_i, & \text{if~~$i=0$}. \\
        \cos{(\omega_i t + \varphi_i)}, & \text{if~~$1\leq i \leq l$},
      \end{cases}
  \end{equation}
  where $\omega_i$ and $\varphi_i$ are trainable parameters.
  We append the time embedding to the raw features of each interaction, denoted by $\vctr{\hat{e}}_{uv}^{k}=(\tau({t^k_{uv}}), \vctr{f}^k_{uv})$.
  The enhanced edge embedding can therefore be formulated as:
  \begin{align}
    \vctr{\tilde{e}}_{uv}=Enc_{i}([\vctr{\hat{e}}_{uv}^1, ..., \vctr{\hat{e}}_{uv}^{S_{uv}}]).
   \label{eq:seq-repr-t2v}
  \end{align}
  The time embedding inherits some good properties of Random Fourier Feartures (RFF) \cite{rahimi2008random}, which enables GTEA to capture more fine-grained temporal behaviours, e.g., time periodicity.
  Detailed analyses refer to Appendix 2.2.

  \subsection{Representation Learning with Temporal Edge Aggregation}
  
  \textbf{Sparsity-Inducing Attention for Neighbors Filtering}~
  Common GNN models learn mutual relationships by iteratively aggregating neighborhood information of a target node.
  To distinguish important nodes in aggregations, dense attention \cite{velivckovic2017graph} is usually applied to calculate an attentive weight for each neighbor.
  However, in real-world tasks for TIGs, e.g., anomaly detection in mobile payment networks, the target node can interact with a large number of neighbors but only few of them are of interest.
  In this case, considerable noisy neighbors, who are assigned small but non-zero attentive weights, can overwhelm the few important, which degrades the representative power of the learned node embeddings.

  With this concern, we propose a sparse attention strategy, motived by \cite{martins2016softmax}, for GTEA to enhance neighbor aggregations.
  The operations are formulated as:
  \begin{align}
    \tilde{\alpha}_{uv}&=\vctr{a}^\intercal \vctr{h}_{uv},~~ \vctr{h}_{uv}=Enc_{a}([\vctr{\hat{e}}_{uv}^1, ..., \vctr{\hat{e}}_{uv}^{S_{uv}}]),\\
    \boldsymbol{\alpha}_{u:}&=\text{Sparse}(\tilde{\boldsymbol{\alpha}}_{u:}),
    \label{equ:sparse-att}  
  \end{align}
  where $Enc_{a}(\cdot)$ maps the interaction sequence of $(u, v)$ to the hidden space, $\vctr{a}$ is a trainable weight vector, $\boldsymbol{\alpha}_{u:}$ is the attentive weight vector for all neighbors of node $u$.
  In experiments, we align $Enc_{a}(\cdot)$ with $Enc_{i}(\cdot)$ but keep independent parameters.
  $\text{Sparse}(\cdot)$ is a sparsification operator (details refer to Appendix 2.3).
  The key idea of $\text{Sparse}(\cdot)$ is to truncate the input by a dynamic threshold. It induces GTEA to learn sparse but normalized attentive weights  $\boldsymbol{\alpha}_{u:}$ for neighbors.
  This forces the model to distinguish the few important neighbors and discard the trivial mass, which refines the information for neighbor aggregations.\\
  \\
  \textbf{Neighbors Aggregation with Temporal Edge Embeddings}~
  With the sparse attention mechanism, neighbors can be selectively aggregated.
  In addition, we incorporate learned edge embeddings into aggregation.
  But instead of simply concatenating node and edge embeddings as the input to the aggregator \cite{xu2020inductive,zhang2020learning}, we propose a new method to correlate node and edge information:
  \begin{align}
    \vctr{z}_{\mathcal{N}(u)}^{(l)} &= \sum_{v \in \mathcal{N}(u)}\alpha_{uv}\mlp_1([\vctr{z}_v^{(l-1)} || \vctr{\tilde{e}}_{uv}]), \label{nbr-repr}\\
    \vctr{z}_u^{(l)} &= \mlp_2([\vctr{z}_u^{(l-1)} || \vctr{z}_{\mathcal{N}(u)}^{(l)}]), \label{node-emb}
  \end{align}
  where $\vctr{z}_u^{(l-1)}$ is the node embedding of $u$ in the $(l-1)$-th layer,  $\mathcal{N}(u)$ is the neighbors set of $u$ and $\text{MLP}_{1}(\cdot)$ and $\text{MLP}_{2}(\cdot)$ are multi-layer perceptrons (MLP), which enables GTEA to fuse features from different latent spaces.
  The introduction of edge embeddings forces GTEA to correlate both temporal and topological dependencies in a TIG and therefore learn more comprehensive and discriminative node representations.
  We prove that Equ \ref{nbr-repr} has more powerful representation ability than the naive concatenation operator.
  Details refer to Appendix 2.4.

  \subsection{Model Training for Different Graph-related Tasks}
  
  By iteratively stacking $L$ GNN layers, GTEA can generate node embeddings with high-level semantics for downstream tasks.
  In this work, we focus on two tasks: node classifications and future link predictions.
  For node classifications, the category probability vector is computed based on:
  \begin{align}
    \vctr{y}_u = \text{Softmax}(\mlp_{3}(\vctr{z}_{u}^{(L)})).
  \end{align}
  For future link predictions, we predict the probability of a future link between $u$ and $v$ by:
  \begin{align}
    y_{uv} = \text{Sigmoid}({\vctr{z}_{u}^{(L)}}^\intercal \vctr{z}_{v}^{(L)}).
  \end{align}
  GTEA can then be trained in an end-to-end manner by cross-entropy loss.
  The steps to train GTEA are summarized in Algorithm 2 in the appendix.
  Different from some previous works \cite{grover2016node2vec,zhang2020learning}, GTEA does not need to maintain a memory to update embeddings, which enables it to work for real-world inductive tasks.  

  \section{Experiments}

  \subsection{Experimental Setup}


  \textbf{Datasets}~
  In our experiments, we formulate node classification as a task to identify illicit users/ nodes.
  We evaluate GTEA on a payment dataset, denoted as Mobile-Pay, which is provided by a major mobile payment provider.
  We additionally assess GTEA on two Ethereum phishing datasets \footnote[2]{\scriptsize{Raw data: https://www.kaggle.com/xblock/ethereum-phishing-transaction-network}} with different scales, denoted as Phish-L(arge) and Phish-S(mall).
  For future link predictions, we use Wikipedia and Reddit datasets \cite{kumar2019predicting} evaluation.
  \textit{Note that we align all other four datasets' setting with Mobile-Pay to follow the product scenarios.
  Thus, our results may not be directly compared with the numbers reported in other works.}
  \textbf{More details and statistics of dataset splits refer to Appendix 3.1.}\\
  \\
  \textbf{Compared Methods}~
  We compare GTEA with different methods, including GNN baselines (GCN \cite{kipf2016semi}, GraphSAGE \cite{hamilton2017inductive} and GAT \cite{velivckovic2017graph}), edge-feature-involved methods (ECConv \cite{simonovsky2017dynamic} and EGNN\cite{gong2019exploiting}) and a state-of-the-art temporal graph learning model (TGAT \cite{xu2020inductive}).
  We also implement a GTEA variant, denoted as GTEA$_{\text{HE}}$, by replacing the learned edge embedding with handcrafted edge features.
  \textit{Note that we mainly focus on the methods applicable to Mobile-Pay, i.e., can learn the overall status of all past interactions.
  Therefore, models cannot fit the product environments, e.g, transductive \cite{nguyen2018continuous,zhang2020learning} or event-by-event learning methods \cite{kumar2019predicting,rossi2020temporal}, are not included.}
  More details refer to Appendix 3.2.\\
  \\
  \textbf{Implementations}~
  We implement two variants of GTEA (regarding the sequence model) by LSTM and Transformer, which are denoted as GTEA$_{\text{L}}$ and GTEA$_{\text{TX}}$, respectively.
  We use `+T' to mark variants enhanced by the time encoder.
  All hyperparameters are tuned through grid-search on the validation set and we report the best \textbf{accuracy} and \textbf{Macro-F1} on the test set.
  \textbf{More implementation details can be found in Appendix 3.3.}

  \subsection{Experimental Results of Overall Performance}


  \begin{table}[t]
    \begin{center}
      \caption{Experimental Results of Node Classifications and Future Link Predictions}
      \label{tab:exp-results}
      \resizebox{\columnwidth}{!}{%
      \begin{tabular}{c|cc|cc|cc|cc|cc} 
        \hline
        \multicolumn{1}{c|}{Tasks} & \multicolumn{6}{c|}{Node Classifications}  & \multicolumn{4}{c}{Future Link Predictions} \\
        \hline
        \multicolumn{1}{c|}{Datasets} & 
        \multicolumn{2}{c}{Mobile-Pay} &
        \multicolumn{2}{c}{Phish-S} &
        \multicolumn{2}{c|}{Phish-L} &	   
        \multicolumn{2}{c}{Wikipedia} &
        \multicolumn{2}{c}{Reddit} \\
        \hline
        Model & Acc & F1 &Acc & F1 &Acc & F1 & Acc & F1 & Acc & F1 \\
        \hline
        GCN & 0.7481 & 0.7480 & 0.9077 & 0.9077 & 0.9298 & 0.8683 & 0.6472 & 0.6259 & 0.5369 & 0.4285\\
        GraphSAGE & 0.7474 & 0.7472 & 0.9405 & 0.9405 & 0.9753 &0.9569 & 0.5986 & 0.5953 & 0.6424 & 0.6334\\
        GAT & 0.7265 & 0.7264 & 0.9405 & 0.9405 & 0.9631 & 0.9375 & 0.6167 & 0.5975 & 0.6396 & 0.6252\\  
        \hline
        ECConv & 0.7399 & 0.7399 & 0.9554 & 0.9559 & 0.9700 & 0.9480 & 0.6426 & 0.6424 &0.6232 & 0.6219\\
        EGNN & 0.7549 & 0.7538 &0.9479 & 0.9477 & 0.9659 & 0.9393 & 0.6401 & 0.6259 & 0.5484 & 0.4406\\
        \hline
        TGAT & 0.7212 & 0.7212 & 0.9673 & 0.9673 & 0.9740 & 0.9559 & 0.7253 & 0.7256 &0.8418 & 0.8414 \\
        \hline
        GTEA$_{\text{HE}}$ & 0.7519 & 0.7516 & 0.9673 & 0.9673 & 0.9777 & 0.9615 & 0.6169 & 0.6123 & 0.6515 & 0.6495\\
        \hdashline
        GTEA$_{\text{L}}$ & 0.7848 & 0.7847 & 0.9836 & 0.9836 & \textcolor{red}{\textbf{0.9805}} & \textcolor{red}{\textbf{0.9668}} & 0.7988 & 0.7981 &0.8809 & 0.8807 \\
        GTEA$_{\text{L}}$+T & \textcolor{red}{\textbf{0.7990}} & \textcolor{red}{\textbf{0.7990}} & 0.9777 & 0.9777 &0.9789 &0.9640 & \textcolor{red}{\textbf{0.8149}} & \textcolor{red}{\textbf{0.8145}} &0.885 & 0.8849 \\
        GTEA$_{\text{TX}}$ & 0.7676 & 0.7670 & \textcolor{red}{\textbf{0.9851}} & \textcolor{red}{\textbf{0.9851}} & 0.9801 & 0.9658 & 0.7841 & 0.7832 & \textcolor{red}{\textbf{0.8865}} & \textcolor{red}{\textbf{0.8864}} \\
        GTEA$_{\text{TX}}$+T & 0.7758 & 0.7758 & 0.9792 & 0.9792 & 0.9769 & 0.9603 & 0.7869 & 0.7864 & 0.8747 & 0.8746\\    
        \bottomrule
      \end{tabular}
      }
    \end{center}
  \end{table}

  \textbf{Node Classifications}~
  Table \ref{tab:exp-results} shows the results for node classifications.
  We can clearly observe that variants of GTEA consistently outperform all other models. 
  We owe such superiority to the edge embedding module and the joint integration of both topological and temporal information.
  With these designs, GTEA can model the interaction dynamics for node pairs, and therefore be more effective to capture discriminative behavior patterns of a node.
  An evidence is that GTEA performs much better in the Mobile-Pay than that of others.
  This may due to that actions of phishing is mostly naive and instantaneous, while illicit payment interactions are associated with more complex temporal patterns, which can be captured more effectively by temporal modeling.
  Even though, GTEA still dominates other competitors over all datasets, which demonstrates its effectiveness.\\
  \\
  \textbf{Future Link Predictions}~
  Results are shown in the right of Table \ref{tab:exp-results}.
  It can be observed that GTEA achieves the best performance in this task.
  Note that TGAT and GTEA perform much better than other competitors that do not incorporate temporal information, which shows the importance of temporal modeling.
  Even though, GTEA still outperforms TGAT by a large margin.
  This is because TGAT mixes all interactions from different neighbors of a target node in temporal modeling, which is hard to distinguish the interactions from the same instance.
  In contrast, we adopt a pairwise interaction modeling scheme, which elaborately exploits the relationship patterns for each neighbor.
  This endows GTEA the power to learn connection features for node embeddings and therefore is more effective for the future link prediction task.

  \subsection{Experiments Analyses}

  \begin{figure*}[tb]
    \centering
    \subfigure{
      \includegraphics[scale=0.4]{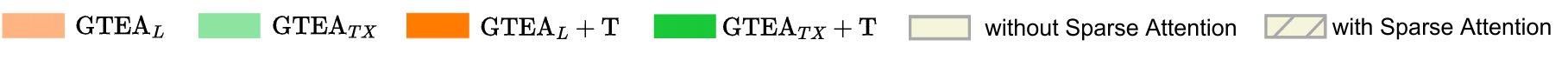}
    }
    \addtocounter{subfigure}{-1}
  
    \subfigure[Phish-S Acc]{
      \label{bar:accuracy:sparsemax:eth:phish2}
      \graphicspath{{charts/}}
      \def\svgwidth{0.20\textwidth}
      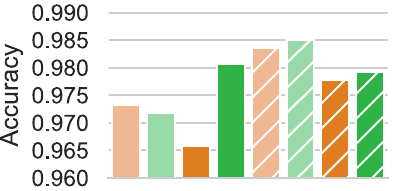
    }
    \subfigure[Phish-S F1]{
      \label{bar:f1:sparsemax:eth:phish2}
      \graphicspath{{charts/}}
      \def\svgwidth{0.20\textwidth}
      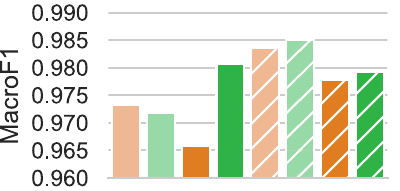
    }
    \subfigure[Mobile-Pay Acc]{
      \label{bar:accuracy:sparsemax:tencent}
      \graphicspath{{charts/}}
      \def\svgwidth{0.20\textwidth}
      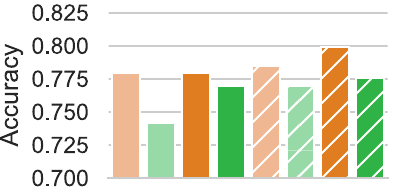
    }
    \subfigure[Mobile-Pay F1]{
      \label{bar:f1:sparsemax:tencent}
      \graphicspath{{charts/}}
      \def\svgwidth{0.20\textwidth}
      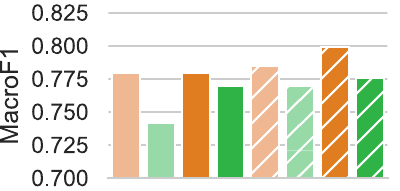
    }
      
    \caption{Effects of the sparse attention mechanism on Phish-S and Mobile-Pay datasets.}
    \label{bar:ab:sparsemax}
  \end{figure*}

  \begin{figure}[t]
    \centering
    \begin{minipage}{0.49\textwidth}
       \centering
       \includegraphics[scale=0.21]{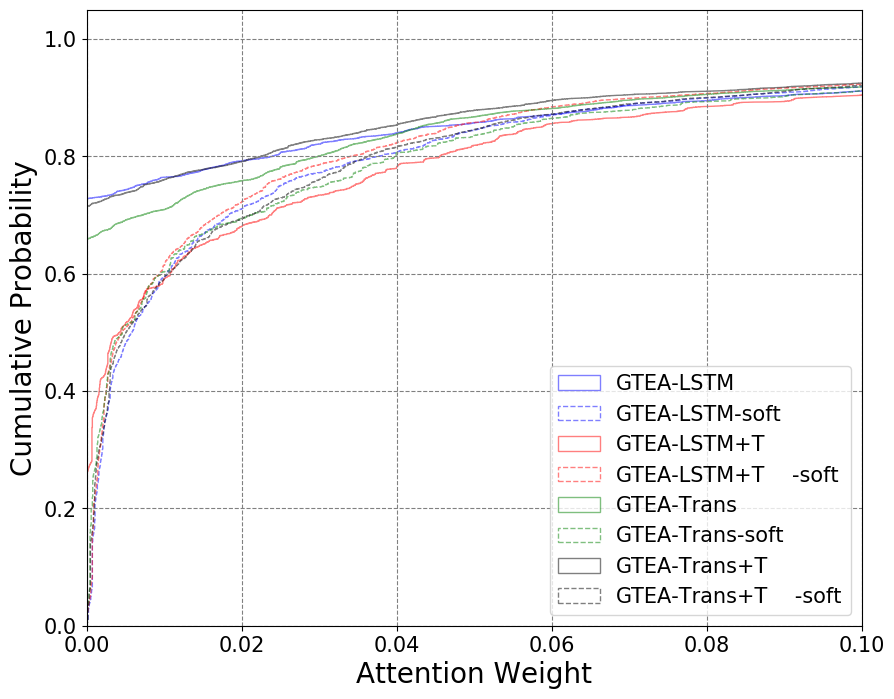}
      \label{fig:ab:attndist:phish}
    \end{minipage}
    \begin{minipage}{0.49\textwidth}
       \centering
       \includegraphics[scale=0.21]{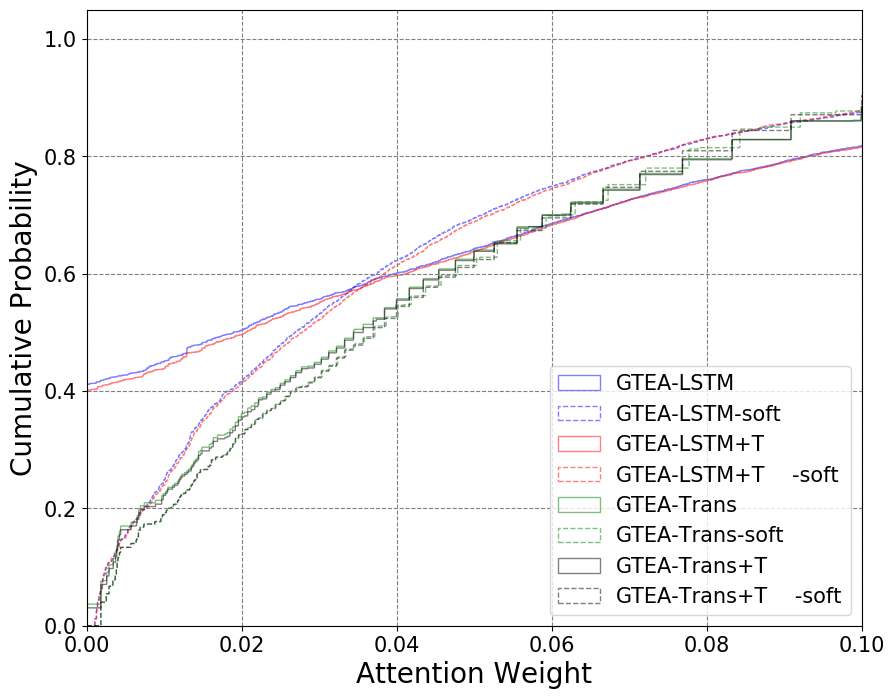}
        \label{fig:ab:attndist:payment}
    \end{minipage}
  
    \caption{Distributions of attention weights (with/without $\text{Sparse}(\cdot)$) on Phish-S (left) and Mobile-Pay (right). The variant with dense attention is denote by ``soft''.}
    \label{fig:ab:attndist}
  \end{figure}

  \textbf{Effects of Edge Features/Embeddings}~
  From Table \ref{tab:exp-results}, we observe that models incorporating edge features mostly perform better than those who learn only from node attributes.
  It is not surprised as edge features carry rich semantics about connections and interactions.
  Information such as user behaviors and potential relations can be mined from edges.
  We also notice that GTEA performs much better than other edge-feature-involved models.
  This is because high-level semantics is encoded into the embedding, which extracts critical information for other modules.
  The additional MLP module introduced in the aggregation process (Equ \ref{nbr-repr}) also helps GTEA to align inputs from different feature domains, which enables it to learn discriminative embeddings more efficiently.\\
  \\
  \textbf{Effects of the Temporal Dynamics Modeling}~
  An advantage of GTEA is to model the temporal dynamics for TIGs, where Table \ref{tab:exp-results} shows the benefit.
  Compared with GTEA$_{\text{HE}}$, which substitutes the learned temporal edge embedding by handcrafted edge features, all GTEA variants perform much better.
  It implies that modeling the temporal dynamics of an interaction sequence is critical for analyzing TIGs.
  The great performance improvement (over GTEA$_{\text{HE}}$) also demonstrates the effectiveness of the temporal dynamics modeling scheme of GTEA.
  In addition, we observe that the time encoder works better in GTEA's LSTM variants than transformer variants.
  We speculate that this is because the function of the time encoder partially overlaps with the position encoder in transformer.
  Instead, vanilla LSTM doesn't encode the position information and therefore benefit much by introducing the time encoder.
  However, when it comes to an environment with more complex and diverse temporal behaviors, e.g., the Mobile-Pay dataset, the time-encoder can be more powerful to capture irregular but discriminative patterns, such as different periodicities, for each interaction sequence, which enhances the representation ability of all GTEA variants.\\
  \\
  \textbf{Effects of the Sparse Attention Aggregation}~
  We conduct additional experiments on Phish-S and Mobile-Pay datasets to validate the effectiveness of the sparse attention mechanism.
  Specifically, we replace the $\text{Sparse}(\cdot)$ operator in Equ \ref{equ:sparse-att} by the Softmax function, which generates dense attentive weights.
  Quantitative results are shown in Fig. \ref{bar:ab:sparsemax}.
  In most cases, models with $\text{Sparse}(\cdot)$ achieve a better performance.
  This is reasonable as redundant and noisy signals of irrelevant neighbors are discarded in aggregation, which encourages the model to yield discriminative node embeddings.
  We additionally visualize the attention weights' distributions of the sparse and dense attention mechanisms, as shown in Fig. \ref{fig:ab:attndist}.
  With dense attention, neighbors with small attentive weights, e.g., $0.1$ in our cases, account for over $80\%$ of all.
  In contrast, for $\text{Sparse}(\cdot)$, around $70\%$ of neighbors are truncated (attentive weights are zeroed) in Phish-S, while $40\%$ in Mobile-Pay.
  Such quantitative and qualitative results demonstrates that noises are substantially suppressed, which explains the effectiveness of GTEA.

  \section{Conclusions}

  In this paper, we propose GTEA for inductive representation learning on Temporal Interaction Graphs (TIGs).
  Different from previous works, GTEA learns an edge embedding for temporal interactions between each pair of adjacent nodes by adopting an enhanced sequence model.
  By incorporating the learned edge embeddings into the aggregation of a GNN, which is driven by a sparse attention mechanism, GTEA is encouraged to exploit both temporal and topological dependencies in TIGs.
  As a general framework, GTEA is evaluated on different graph-related tasks and extensive experimental results show its effectiveness.\\
  \\
  \textbf{Acknowledgements}
  This research is supported in part by the Innovation and Technology Committee of HKSAR under the project\#ITS/244/16, the CUHK MobiTeC R\&D Fund and a gift from Tencent.

%
%
%

\bibliographystyle{splncs04}
\bibliography{PAKDD23_GTEA}

\end{document}


\title{Appendices for:\\
GTEA: Inductive Representation Learning on Temporal Interaction Graphs via Temporal Edge Aggregation}
\titlerunning{Inductive Learning for Temporal Interaction Graphs with GTEA}
%
\author{Siyue Xie \inst{1}$^\star$, Yiming Li \inst{1}\thanks{Equal contributions.}, Da Sun Handason Tam \inst{1} \and
Xiaxin Liu \inst{2}, Qiufang Ying \inst{2} \and
Wing Cheong Lau \inst{1}, Dah Ming Chiu \inst{1} \and
Shouzhi Chen \inst{2}
}
%
\authorrunning{Siyue Xie, Yiming Li, et al.}
%

\institute{
  The Chinese University of Hong Kong \and Tencent Technology Co.Ltd
}

\maketitle              

\section{More Discussions and Comparisons on Related Works}

\subsection{Comparisons on Temporal Inductive Models}

In real-world environments, the topology of a TIG usually changes as time goes.
For example, in an online social network, not only existing users interact with each other, but also new users be added and they will communicate with others.
Such situations require that models adaptive to real-world product scenarios should be with an inductive nature and capable of dealing with temporal dynamics.

Many previous methods attempt to learn from temporal graphs but most of them \cite{nguyen2018continuous,zhang2020learning} can only work for transductive tasks.
But there still some recent works attempt to inductively learn representations for temporal graphs.
TDGNN \cite{qu2020continuous} inherits the working pipeline of GraphSAGE \cite{hamilton2017inductive} and extend it to temporal graphs by incorporating the time information of each interaction.
The object of each interaction of a target node is treated as an independent neighbor.
Neighbors aggregation can therefore be applied to all interactions, accompanied with an time-exponential decreasing weight to attenuate the importance of a past event.
The framework of TGAT \cite{xu2020inductive} is similar to TDGNN but instead introduce a self-attention mechanism to weight different interactions/ neighbors for aggregation.
In addition, a time-encoding method is incorporated to embed time information.
CAWs \cite{wang2021inductive} proposes to utilize causal anonymous walks on temporal graphs to generate node embeddings, which helps mine some local motifs of the graph structure.

However, there is still a common drawback of the aforementioned inductive methods: the identity information of interactions is ignored or even purposefully removed.
However, in TIGs, we argue that the identity of each interaction object can help models learn more discriminative node representations, as the illustrative example we shown in Figure 1 in our main text. 
In CAW \cite{wang2021inductive}, to capture temporal motifs structures, different nodes with the same causal walk orders will result in similar representations.
TDGNN \cite{qu2020continuous} and TGAT \cite{xu2020inductive} lumps all interactions of a target node for aggregation.
Although the self-attention mechanism can help TGAT to highlight some more important interactions, it is unable to distinguish different interactions triggered by the same neighbor.
In contrast, our proposed model, GTEA, learns an embedding to represent a sequence of interactions between a pair of adjacent nodes.
Interactional patterns encoded in an edge embedding are consistently extracted from the same instance, which preserves the discriminative identity information for aggregation.
Experimental results shown in Section 4 in main text also demonstrate the superiority of GTEA over TGAT.

\subsection{Comparisons on Dynamic Embedding Methods}

Dynamic embedding methods are commonly used to learn node representations for TIGs.
Specifically, a node embedding is usually randomly-initialized and the model will hold a memory to store the embeddings (or status related to embeddings) of all nodes.
By observing the interactions happened along time, node embeddings can be dynamically updated.

TigeCMN \cite{zhang2020learning} leverages the recent advances in memory networks to learn temporal interactional features as well as node multi-faceted properties.
MNCI \cite{liu2021inductive} additionally learns a group of community embeddings to model the community influence on node embeddings.
JODIE \cite{kumar2019predicting} focuses on user-item bi-partite graphs and utilizes two independent RNN models to learn node representations by tracking the embedding trajectories.
TGN \cite{rossi2020temporal} additionally introduces a time-encoder and implements a training strategy that allows it to update and learn node embeddings with efficient parallelization.

Although dynamic embedding methods have achieved great success, the drawbacks can also be obvious.
First, since a memory is required to maintain the embedding of existing nodes, some of the methods are hard to generalize or even fail to work when there is new nodes added (i.e., inductive tasks).
Second, the corresponding node embeddings are always updated once an interaction is observed.
The status of a node may change frequently, which is not expected for some downstream tasks, e.g., anomaly detection or phishing node detection, where only few interactions of the entire interaction history carry critical hints.
Such a updating paradigm may fail to capture the most discriminative patterns in the interactions.
Consequently, it is sometimes hard to yield a general representations that can depict the profile of a user, which is very important in some tasks, e.g., to identify the role of a user in a TIG as in our experiments.
Last but not least, the model have to maintain a big and heavy memory to remember and update all nodes it observed.
The condition may get worse or even inapplicable when it comes to a large-scale graph with million or billion nodes.
The above drawbacks are also the reasons why these methods are inapplicable to the dataset we mainly concerned, i.e., the Mobile-Pay dataset.

Compared with the aforementioned models, GTEA is designed to work for inductive tasks (by learning model weights instead of learning node embeddings directly), which easily handles the cases when the graph topology evolves.
Moreover, GTEA models the interactional dynamics by observing the entire interaction history between a pair of nodes.
This helps it mine temporal relational patterns from a global view and more helpful to summarize the status of each node.
Results on different tasks in our experiments also demonstrate the effectiveness of GTEA.

\section{Technical Details of the Proposed Method}

In this section, we provide more technical details of the key modules of GTEA.

\subsection{Interaction Dynamics Modeling with Different Sequence Models}

We implement different GTEA variants with two sequence models, i.e., LSTM and Transformer, for interactional dynamics modeling.
In the following, we specify the details.

\subsubsection{Temporal Dynamics Modeling with LSTM}

LSTM \cite{hochreiter1997long} is implemented to capture the patterns of consecutive interactions between a pair of nodes.
Concretely, let's denote $[\vctr{e}_{uv}^1, ..., \vctr{e}_{uv}^{S_{uv}}]$ as the interaction sequence of edge $(u, v)$, where $S_{uv}$ indicates the total number of events between a specific node pair. 
Note that $S_{uv}$ can be different for different node pairs. 
We therefore adapt the LSTM model to encode the interaction sequence of edge $(u,v)$ as follows (we drop the subscript $_{uv}$ for notational convenience):

\begin{align}
  \vctr{i}[k] &= \sigma\left(\matr{W}_i
  \vctr{e}^k+\matr{U}_i\vctr{h}[k-1]+\vctr{b}_i\right) \\
  \vctr{f}[k] &= \sigma\left(\matr{W}_f\vctr{e}^k+\matr{U}_f\vctr{h}[k-1]+\vctr{b}_f\right) \\
  \overline{\vctr{c}}[k]  &= \tanh\left(\matr{W}_c \vctr{e}^k+\matr{U}_c \vctr{h}[k-1]+\vctr{b}_c\right) \\
  \vctr{c}[k]  &= \vctr{f}[k]\odot\vctr{c}[k-1] + \vctr{i}[k] \odot \overline{\vctr{c}}[k] \\
  \vctr{o}[k] &= \sigma\left(\matr{W}_o \vctr{e}^k+\matr{U}_o \vctr{h}[k-1]+\vctr{b}_o\right) \\
  \vctr{h}[k] &= \vctr{o}[k] \odot\tanh\left(\vctr{c}[k] \right)
\end{align}

\sloppy
$\vctr{e}^k$ represents the $k$-th step's input of the LSTM. 
In GTEA, $\vctr{e}^k$ corresponds to the features of the $k$-th interaction event.
$\vctr{i}[k]$, $\vctr{f}[k]$, and $\vctr{o}[k]$ are state vectors of the input gates, forget gates and output gates respectively.
$\vctr{c}[k]$ is the memory cell and $\vctr{h}[k]$ indicates the hidden state. 
$\sigma(\cdot)$ and $\tanh(\cdot)$ represent the sigmoid and hyperbolic tangent activation functions respectively.
$\matr{W}_i, \matr{W}_f, \matr{W}_c, \matr{W}_o, \matr{U}_i, \matr{U}_f, \matr{U}_c, \matr{U}_o, 
\vctr{b}_i, \vctr{b}_f, \vctr{b}_c, \vctr{b}_o$ are trainable parameters of the network.
We take the last hidden output $\vctr{h}_{uv}[S]$ of the LSTM as the edge embeddings $\vctr{\tilde{e}}$.
Note that $\vctr{e}^k$ will be replaced by $\vctr{\hat{e}}^k$ when the edge features are enhanced by the time-embedding.

\subsubsection{Temporal Dynamics Modeling with Transformer}
Transformer \cite{vaswani2017attention} is another popular architecture for sequence modeling.
The key component of the Transformer architecture is the self-attention mechanism, where a self-attention function maps a query and a set of key-value pairs to an output. 
In particular, given a pair of nodes $(u, v)$, we formulate its interaction sequence as a matrix: $\matr{E}_{uv}=[\vctr{e}^1, ..., \vctr{e}^{S_{uv}}]^\intercal$.
Then the projection of the interaction sequence on the query, key and value space can be computed as:
\begin{align}
  \matr{Q}&=\matr{E}_{uv}\matr{W}_{Q}, \\
  \matr{K}&=\matr{E}_{uv}\matr{W}_{K}, \\
  \matr{V}&=\matr{E}_{uv}\matr{W}_{V}. 
\end{align}
where $\matr{Q, K, V}$ denotes the queries, keys and values matrices, respectively.
$\matr{W}_{Q}, \matr{W}_{K}, \matr{W}_{V}$ are learnable parameters.
Then the ``Scaled Dot-Product Attention'' function can be defined as: 
\begin{align}
  \matr{\tilde{E}}_{uv} = \text{Attention}(\matr{Q,K,V}) = \text{softmax}(\frac{\matr{QK^T}}{\sqrt{D_{E}}})\matr{V}
\end{align}
where $D_{E}$ is the hidden dimensions of a query vector.
Since Transformer implicitly computes the mutual correlations of all interactions and the last interaction empirically reflects the latest relation status of a node pair, it is sufficient to represent the edge embedding $\vctr{\tilde{e}}_{uv}$ by the $S$-th row of the $\matr{\tilde{E}}_{uv}$, i.e., the embedding corresponding to the last interaction.
Note that for time-encoder-enhanced variants, the input of Transformer, i.e., the interaction matrix $\matr{E}_{uv}$, will also be replaced by $\matr{\hat{E}}_{uv}=[\vctr{\hat{e}}^1, ..., \vctr{\hat{e}}^{S_{uv}}]^\intercal$.

In our experiments, we implement a multi-head attention transformer for GTEA.
The final edge embedding is yielded by the mean of each attention head.

\subsection{Additional Analyses on the Time-embedding}

In the interaction sequence of an edge $(u,v)$, the time gap of any two consecutive interactions can vary with great differences.
In order to capture more fine-grained time-related patterns of an interaction sequence, e.g., time periodicity, we follow the ideas of Time2Vec \cite{mehran2019time2vec} (\ttov) to generate a time-embedding to further enhance the edge interaction features.
The key idea of the time-embedding is to map the timestamp of each interaction from a scalar to a vector, which is capable of encoding different time-related properties.

Specifically, we generate a time embedding with $l$ dimensions for any given timestamp $t$ as follows:
\begin{equation}
  \label{eq:t2vec_def}
  \tau(t)[i]=
    \begin{cases}
      \omega_i t + \varphi_i, & \text{if~~$i=0$}. \\
      \cos{(\omega_i t + \varphi_i)}, & \text{if~~$1\leq i \leq l$}.
    \end{cases}
\end{equation}
where $\omega_0', ... \omega_l'$ and $\varphi_0', ..., \varphi_l'$ are trainable parameters.

\sloppy
On the theoretical side, $\tau(t)$ can be viewed as an extension of Random Fourier Features (RFF) \cite{rahimi2008random}. 
If we define: 
\begin{equation}
  \mathbf{\gamma}(t) = \sqrt{\frac{2}{k}}[\cos(\omega_1't + \varphi_1'), ... ,\cos(\omega_k't + \varphi_k')]^\intercal
\end{equation}
where $\omega_1', ... \omega_k'$ are i.i.d samples from some probability distribution $p(\omega)$ and 
$\varphi_1', ..., \varphi_k'$ are i.i.d samples from the uniform distribution on $[0, 2\pi]$.
Then, as a consequence of the Bochner's theorem from harmonic analysis, 
it can be shown that $\mathbb{E}[\mathbf{\gamma}(t_1)^\intercal \mathbf{\gamma}(t_2)] = \phi(t_1, t_2)$ 
for some positive definite shift-invariant kernel $\phi(t_1, t_2) = \phi(t_1 - t_2)$. 
Thus, the time-embedding can be regarded as RFF with tunable phase-shifts and frequencies. 

Empirically, the $i$-th element (except when $i=0$) of $\tau(t)$ is defined by a harmonic function, whose value recurs with a period of $\frac{2\pi}{\omega_{i}}$.
Therefore, $\tau(t)$ has the capability to encode the periodicity of a sequence of timestamps.
For example, if $t$ is measured by `hours' and some interactions between two specific nodes always happen at the same time of a day, then the time-embedding can encode the daily recurrence by making $\omega_{i} = \frac{2\pi}{24}=\frac{\pi}{12}$.
In other words, the interactions occur at a specific time will share the same embedding value.
In our experiment, we let GTEA to learn such time-related patterns in an end-to-end manner based on different downstream tasks. 

It is worth noting that sinusoidal activation is also found to be suitable for representing complex natural signals \cite{sitzmann2019siren} with high precision.
In addition, sinusoidal functions with fixed frequencies and phase-shifts have also been used in the Transformer model \cite{vaswani2017attention} as positional encodings.
However, it has been shown that learning the frequencies and phase-shifts as in $\ttov$ rather than fixing them can achieve better performance.

\subsection{Details of the Sparsity-Inducing Attention}

\begin{algorithm}[tb] 
  \caption{Sparsemax Transformation} 
  \label{alg:sparsemax} 
  \begin{algorithmic}[1] 
    \STATE \textbf{Input:} $\boldsymbol{z}$
    
\STATE Sort $\boldsymbol{z}$ as $z_{(1)}\geq ... \geq z_{(K)} $
\STATE Find $k(\vctr{z}):=$max$\{k \in [K] | 1+kz_{(k)} \geq \sum_{j \leq k} z_{(j)}\}$
\STATE Define $\gamma(\vctr{z})=\frac{\sum_{j \leq k} z_{(j)}-1}{k(\vctr{z})}$
    \STATE \textbf{Output:} $\boldsymbol{p}$ s.t. $p_i=[z_i-\gamma(\boldsymbol{z})]_{+}$
    
  \end{algorithmic}   
\end{algorithm}

In GTEA, we propose to learn sparse attentive weights for neighborhood aggregation.
The process can be formulated as follows:
\begin{align}
  \tilde{\alpha}_{uv}&=\vctr{a}^\intercal \vctr{h}_{uv},~~ \vctr{h}_{uv}=Enc_{a}([\vctr{\hat{e}}_{uv}^1, ..., \vctr{\hat{e}}_{uv}^{S_{uv}}]),\\
  \boldsymbol{\alpha}_{u:}&=\text{Sparse}(\tilde{\boldsymbol{\alpha}}_{u:}),
  \label{equ:sparse-att}  
\end{align}
In particular, we learn an attention embedding $\vctr{h}_{uv}$ for each interaction sequence between the target node and a neighboring node.
We then map the attention embedding into a scalar by computing the inner-product of it with a learnable vector $\vctr{a}$, which yields an attention measurement for each edge.
We finally apply a sparsification operator, $\text{Sparse}(\cdot)$ to sparsify the attentive weights, which essentially discard some noisy neighbors for the neighborhood aggregation.
In GTEA, we adopt Sparsemax \cite{martins2016softmax} as our sparsification operator.
Algorithm \ref{alg:sparsemax} shows the detailed sparsification steps.
Note that the sparsification operator $\text{Sparse}(\cdot)$ is not unique and can be replaced by any gradient-continuous operators if suitable.

\subsection{Additional Analyses on Temporal Edge Aggregation}

\label{sect:NodeReprWEdgeAgg}
To obtain an expressive node representation, we integrate edge embeddings with neighbor node embeddings. 
To jointly learn node and edge information, \cite{xu2020inductive,zhang2020learning} simply concatenates the node embedding with the edge embedding. 

However, \textbf{we argue that: the operation of simply concatenating node and edge embeddings is inferior for jointly learning instance-aware features for neighborhood aggregations.} 
We empirically show it as as follows: as shown in Equation~(\ref{eq:cat-lin}), when concatenating node and edge embeddings, we can reformulate the aggregation function by splitting the transformation weight matrix $W$ into two independent matrices $W=[W_1, W_2]$: 
\begin{align}
  \vctr{z}_{\mathcal{N}(u)}^{(l)} &= \sum_{v \in \mathcal{N}(u)}\mathbf{W}([\vctr{z}_v^{(l-1)} || \vctr{\tilde{e}}_{uv}]]) \notag\\
 &=\mathbf{W}_1\sum_{v \in \mathcal{N}(u)}\vctr{z}_v^{(l-1)} + \mathbf{W}_2\sum_{v \in \mathcal{N}(u)} \vctr{\tilde{e}}_{uv} \label{eq:cat-lin} 
\end{align}
Equation~(\ref{eq:cat-lin}) shows that: the neighboring node embeddings and the corresponding edge embeddings are essentially summed independently \cite{li2019gcn}.
Node embeddings are fully isolated with their corresponding edge embeddings, which makes the model agnostic to the identity (node) of an edge embedding and therefore messes up the interactional information from all neighbors.
In other words, it is not a good choice to jointly learn identity-aware features by applying naive concatenation of node and edge embeddings. \\[4pt]

To get rid of such a drawback, we propose to introduce additional projection functions, e.g., a MLP, before aggregation to address the problem:
\begin{align}
  \vctr{z}_{\mathcal{N}(u)}^{(l)} &= \sum_{v \in \mathcal{N}(u)}\alpha_{uv}\mlp_1([\vctr{z}_v^{(l-1)} || \vctr{\tilde{e}}_{uv}]) \label{nbr-repr}\\
  \vctr{z}_u^{(l)} &= \mlp_2([\vctr{z}_u^{(l-1)} || \vctr{z}_{\mathcal{N}(u)}^{(l)}]) \label{node-emb}
\end{align}
where $\mlp_1$ and $\mlp_2$ are two independent MLP models. 
Equation~(\ref{nbr-repr}) applies attention weighted sum to aggregate both the node embeddings $\vctr{z}_u^{(l-1)}$ and the associated edges embedding vector $\vctr{\tilde{e}}_{uv}$ of its neighbors at layer $(l-1)$, where $\alpha_{uv}$ is derived from the sparse attention mechanism.
We can show that:\\[8pt]
\textbf{Theorem:}~ The representation ability of the aggregation operator defined in Equation~(\ref{nbr-repr}) is strictly more powerful than that of Equation~(\ref{eq:cat-lin}).\\[8pt]
\textit{Proof:}~ Let $\alpha_{uv}=\frac{1}{|\mathcal{N}(u)|}$ and degrade $\mlp_1$ to a single-layer fully-connected neural networks without non-linear activations, which can be denoted as $\mlp_1(\mathbf{x}) = |\mathcal{N}(u)| \mathbf{W}\mathbf{x}$. 
Then Equation~(\ref{nbr-repr}) is degraded to the same form as Equation~(\ref{eq:cat-lin}).
Therefore, the representation ability of Equation~(\ref{nbr-repr}) is naturally more powerful than that of Equation~(\ref{eq:cat-lin}). \\[4pt]

$\mlp_1$ projects the concatenated embedding (of node and edge from the same instance/ neighbor) to a hidden representation space, which forces the model to correlate the node and its corresponding edge information before aggregation.
Such an operator empirically preserves the identity information for TIGs. 

After computing the neighboring feature vector $\vctr{z}_{\mathcal{N}(u)}^{(l)}$, 
we concatenate it with the target node's embedding in the previous layer $\vctr{z}_u^{(l-1)}$ and feed it through another MLP, i.e., $\mlp_2$, as in Equation~(\ref{node-emb}). 
By combining the aggregation module and the pairwise temporal sequence model, We can recursively stack the framework at a depth of $L$ layers to learn node representations with high-level semantics.

\section{Supplementary Materials of Experiments}

In our experiments, we evaluate our proposed model by the node classification task on a real-world industrial dataset (denoted as Mobile-Pay), provided by a major mobile payment provider.
The classification problem is formulated as anomaly detection, i.e., a binary classification task.
We additionally apply GTEA on two other real-world datasets to evaluate its performance for node classifications.
Furthermore, to evaluate the generalization ability of representation performance, we extend GTEA to the task of future link predictions.
\textbf{\textit{Note that we align the experimental setups of other datasets with Mobile-Pay in order to follow the real-world product scenarios.
Therefore, some previous methods \cite{nguyen2018continuous,zhang2020learning,kumar2019predicting,rossi2020temporal} that cannot fit such settings are not implemented for comparison.
Since our experimental settings are different from some other works \cite{kumar2019predicting,rossi2020temporal}, their reported results on the same dataset are unsuitable to directly compare with our results.}}

\subsection{Datasets}

\begin{table}[tb]
  \begin{center}
    \caption{Dataset Summary}    
    \label{tab_DS}
    \begin{tabular}{c|c|c|c|c} 
    \hline
    \hline
      Dataset & \# Nodes & \# Edges & \# Interactions  & \# Labeled Nodes (\# Samples) \\
      \hline
      Mobile-Pay & 2,143,844 & 4,568,936 & 21,326,122 & 6,688\\
      \hline
      Phish-S  & 1,329,729 & 2,161,573& 6,794,521   & 3,360 \\
      \hline
	  Phish-L  & 2,973,489 & 5,355,155& 184,398,820 &6,165\\  
      \hline
      Wikipedia & 9,227 & 20,922 & 157,474 & 26,678 \\
      \hline
      Reddit & 10,984 & 103,956 & 672,447 & 156,314 \\
      \hline
    \end{tabular}
  \end{center}
\end{table}

We conduct node classifications on three datasets: Mobile-Pay, Phish-S and Phish-L.
For future link predictions, we evaluate GTEA on the Wikipedia and Reddit datasets.
In the following, we introduce each of these datasets and elaborate the experimental settings.
A brief statistics of these five datasets are summarized in Table \ref{tab_DS}.


\subsubsection{Mobile Payment Network Dataset}
With the rapid growth of mobile payment services, illicit users bring critical challenges to both service provides and regulators. 
We therefore collect a dataset, named as \textit{Mobile-Pay}, from a major mobile payment provider, with 3347 users labeled as ``illicit'' and 3341 normal users, to form a TIG for anomaly detection. 
In the graph, a node indicates a user.
Transactions between two users are represented as an edge, which corresponds to a sequence of interaction events. 
Each node is associated with some static attributes to describe the user, while each interaction of an edge records some basic information of a transaction, e.g., amount and time.
After filtering out edges with few transaction counts and nodes with extremely large degrees, the resultant graph has 2,143,844 nodes and 21,326,122 edges.
All user-related information is anonymized for privacy security concerns.
Node and edge features, such as transaction amount, are normalized before experiments.
\textbf{The goal of GTEA on this dataset is to identify the role of a user (whether it is an illicit user or normal user), given the history of all transaction interactions in the past.}

\subsubsection{Phishing Dataset} 
We collect two different Ethereum Phishing Datasets provided by \cite{xblockEthereum}, referred as \textit{Phish-S} (`S' stands for `small') and \textit{Phish-L} (`L' stands for `large') respectively\footnote{\scriptsize{Raw data: https://www.kaggle.com/xblock/ethereum-phishing-transaction-network}}. 
\textbf{The goal on these two datasets is to determine whether an account in the Ethereum network is a phishing account.}
To build the graph, each account is viewed as a node and transaction sequences between two nodes indicate an edge.
For each user account, we compute some statistical features as node attributes based on the transaction records it involved.
In \textit{Phish-S}, we obtain 1,660 phishing nodes and 1,700 normal nodes.
In \textit{Phish-L}, 1,165 nodes are annotated as phishing users and we randomly select 5,000 nodes from the rest as normal users, which sums up to 6,665 labeled nodes.

\subsubsection{Wikipedia Dataset}
Wikipedia is the world's most widely used online encyclopedia. 
\cite{kumar2019predicting} collects the editing events in Wikipedia from the internet\footnote{http://snap.stanford.edu/jodie/wikipedia.csv} and form the \textit{Wikipedia} dataset.
Each event (interaction) records the editing content and timestamp of a user on a wiki-entry.
The text editing content is vectorized as an embedding in advance, which is used as the raw features of each interaction in an edge.
As for nodes, we use a randomly initialized vector to represent node attributes.
The dataset contains 8,277 users and 1,000 items (wiki-entry). 
There are totally 157,474 interaction between users and items. 
We sort the interactions based on their timestamps and take the first 50\% of interactions to construct the graph, where users or items are nodes and an interaction sequence of a user-item pair induces an edge.
\textbf{ 
The goal of GTEA on this dataset is to determine whether a user will interact with an item at \textit{any time of the future}, given all the interaction history in the constructed graph.}
Therefore, we randomly split the remaining interactions into a training set (60\%), validation set (20\%) and test set (20\%). 
Node pairs found interacted in the train/ validation/ test set are regarded as positive samples.
To construct negative samples, we randomly pick one item that does not have interactions with the user (of each positive sample) to form a negative user-item pair.
The resultant graph has 9,277 nodes and 20,922 edges with 26,678 (positive and negative) samples in total.    

\subsubsection{Reddit Dataset}
The Reddit dataset is also provided by \cite{kumar2019predicting}, which contains 1,000 users and 984 items with 672,447 interactions among them\footnote{
http://snap.stanford.edu/jodie/reddit.csv}. 
\textbf{
The setting on Reddit is similar to that of \textbf{Wikipedia}.}
The resultant graph has 10,984 nodes and 103,956 edges with 156,314 samples. 
\\
\\
\textbf{On each of the datasets, we apply five-fold cross-validation strategy to evaluate our model.}

\subsection{Implementation Details of Compared Methods and Different GTEA Variants}

To evaluate the performance of GTEA, we compare our model with state-of-the-art methods that are able to handle large-scale graphs. 
Specifically, there are six baseline models and five GTEA variants, which can be categorized as follows: 
\begin{itemize}


\item \textbf{GNNs learned only with node features}. These methods are baseline GNNs that learn node embeddings without edge features or temporal information. 
We take GCN\cite{kipf2016semi}, GraphSAGE \cite{hamilton2017inductive}, GAT \cite{velivckovic2017graph} as the representative models for comparison.
Since these methods cannot utilize time-related information of a TIG, i.e., interaction sequences are not incorporated during training, the learned node embeddings only reflect the topological dependencies.

\item \textbf{Methods incorporate edge features}. This category of models incorporate node features as well as statistical edge features that are manually derived from temporal interactions between each node pair. 
We implement ECConv\cite{simonovsky2017dynamic} and EGNN\cite{gong2019exploiting} as the representatives for comparison. 
We also implement a variant of GTEA named GTEA$_\text{HE}$, which replaces the automatically learned edge-embedding with handcrafted interaction-statistics-based edge features.
The handcrafted edge features include the mean, minimum, maximum and standard deviation amount of feature values of each interaction sequence.

\item \textbf{GNNs with temporal learning}. We also compare GTEA with 
TGAT \cite{xu2020inductive}, a state-of-the-art representation learning scheme for temporal graphs learning.

\item \textbf{GTEA variants}. We implement different GTEA variants regarding the sequence model.
Specifically, the variants equipped with LSTM and Transformer are denoted by GTEA$_\text{L}$ and GTEA$_\text{TX}$ respectively. 
Besides, the variants enhanced by the time encoder are marked with a `+T' suffix, including GTEA$_\text{L}$+T and GTEA$_\text{TX}$+T.

Note that we do not implement dynamic embedding methods for comparisons.
Reasons are elaborated in Appendix 1.2.

\end{itemize}

\subsection{Training Details}

\begin{algorithm}[tb]
    \caption{GTEA Framework (One Forward-Propagation Step)}
    \label{alg:GTEA} 
    \textbf{Input}: $\Graph=(\V, \E)$; 
      node features $x_{u}, \forall u\in\V$;
      temporal edge sequences $\matr{e}_{uv}, \forall (u, v)\in\E$;
      depth of GNN backbone $L$;
      aggregation functions $\mlp_1$ and $\mlp_2$;
      neighborhoods of node $u$: $\mathcal{N}(u)$;
      edge embedding model $Enc_i$ and attention model $Enc_a$;
      trainable attention weight vector $\vctr{a}$.\\
    \textbf{Output}: Node embedding $\vctr{z}_u^{(L)}$
    \begin{algorithmic}[1] 
    \STATE $\vctr{z}_u^{(0)} \leftarrow x_{u}$;
      \FOR{$l = 1, 2, ..., L$}
          \STATE $\vctr{\tilde{e}}_{uv} \leftarrow Enc_{i}(e_{uv}), \forall v\in\mathcal{N}(u)$;
          \STATE $\vctr{h}_{uv} \leftarrow Enc_{a}(e_{uv}), \forall v\in\mathcal{N}(u)$;
          \STATE $\tilde{\alpha}_{uv}\leftarrow \vctr{a}^\intercal\vctr{h}_{uv}$ 
          \STATE $\vctr{\alpha}_{uv} \leftarrow \text{Sparse}_v(\tilde{\alpha}_{u:})$
          \STATE $\vctr{z}_{\mathcal{N}(u)}^{(l)} \leftarrow \sum_{v \in \mathcal{N}(u)}\alpha_{uv}\mlp_1([\vctr{z}_v^{(l-1)} || \vctr{\tilde{e}}_{uv}]])$;
           \STATE $\vctr{z}_u^{(l)} \leftarrow \mlp_2([\vctr{z}_u^{(l-1)} || \vctr{z}_{\mathcal{N}(u)}^{(l)}])$
      \ENDFOR
    \RETURN $\vctr{z}_{u}^{(L)}$; 
    
    \end{algorithmic}
\end{algorithm}

We summarize the algorithm of GTEA in Algorithm \ref{alg:GTEA}.
All the algorithms are implemented using Pytorch \cite{paszke2019pytorch} and DGL v0.5 \cite{wang2019deep}. 
For fair comparison, we tuned all baselines and GTEA variants with different hyperparameter settings. 
We search learning rate in \{0.0001, 0.001, 0.01\}, number of GNN layers in \{1, 2, 3\}, hidden embedding dimension in \{32, 64, 128\} for both nodes and edges. 
For models using LSTM, i.e. our GTEA$_\text{L}$ and GTEA$_\text{L}$+T variants, the number of temporal layers is searched from \{1, 2, 3\}. 
For Transformer-based variants, i.e. GTEA$_\text{TX}$ and GTEA$_\text{TX}$+T, the number of transformer layers is searched among \{1, 2, 3\}, and the number of attention heads is set to 4. 
To make it train on large-scale graphs, we apply mini-batch training with the neighbor sampling strategy \cite{hamilton2017inductive} for all models if applicable. 
We tune the batch size with values in \{32, 64, 128\}, and the neighbors sampling size in \{5, 10, 25\}. 
Adam \cite{kingma2014adam} is used as the optimizer in our training stage. 
For all other models implemented for comparisons, their tuning process is the same as that of GTEA.
We tune the hyperparameters based on the validation set and we report the best accuracy and macro F1 score on the test set.

Our experiments are conducted on a single Linux machine with an AMD Ryzen Threadripper 3970X CPU @ 3.7GHz, 256 GB RAM, and a NVIDIA RTX TITAN GPU with 24GB of RAM.

%
%
%
\bibliographystyle{splncs04}
\bibliography{PAKDD23_GTEA}
%






%% file: charts/barplot_Accuracy_sparsemax_eth_phish2.pdf_tex
\begingroup%
  \makeatletter%
  \providecommand\color[2][]{%
    \errmessage{(Inkscape) Color is used for the text in Inkscape, but the package 'color.sty' is not loaded}%
    \renewcommand\color[2][]{}%
  }%
  \providecommand\transparent[1]{%
    \errmessage{(Inkscape) Transparency is used (non-zero) for the text in Inkscape, but the package 'transparent.sty' is not loaded}%
    \renewcommand\transparent[1]{}%
  }%
  \providecommand\rotatebox[2]{#2}%
  \ifx\svgwidth\undefined%
    \setlength{\unitlength}{192.08655954bp}%
    \ifx\svgscale\undefined%
      \relax%
    \else%
      \setlength{\unitlength}{\unitlength * \real{\svgscale}}%
    \fi%
  \else%
    \setlength{\unitlength}{\svgwidth}%
  \fi%
  \global\let\svgwidth\undefined%
  \global\let\svgscale\undefined%
  \makeatother%
  \begin{picture}(1,0.4763636)%
    \put(0,0){\includegraphics[width=\unitlength,page=1]{barplot_Accuracy_sparsemax_eth_phish2.pdf}}%
  \end{picture}%
\endgroup%

%% file: charts/barplot_MacroF1_sparsemax_eth_phish2.pdf_tex
\begingroup%
  \makeatletter%
  \providecommand\color[2][]{%
    \errmessage{(Inkscape) Color is used for the text in Inkscape, but the package 'color.sty' is not loaded}%
    \renewcommand\color[2][]{}%
  }%
  \providecommand\transparent[1]{%
    \errmessage{(Inkscape) Transparency is used (non-zero) for the text in Inkscape, but the package 'transparent.sty' is not loaded}%
    \renewcommand\transparent[1]{}%
  }%
  \providecommand\rotatebox[2]{#2}%
  \ifx\svgwidth\undefined%
    \setlength{\unitlength}{191.98155954bp}%
    \ifx\svgscale\undefined%
      \relax%
    \else%
      \setlength{\unitlength}{\unitlength * \real{\svgscale}}%
    \fi%
  \else%
    \setlength{\unitlength}{\svgwidth}%
  \fi%
  \global\let\svgwidth\undefined%
  \global\let\svgscale\undefined%
  \makeatother%
  \begin{picture}(1,0.47662414)%
    \put(0,0){\includegraphics[width=\unitlength,page=1]{barplot_MacroF1_sparsemax_eth_phish2.pdf}}%
  \end{picture}%
\endgroup%

%% file: charts/barplot_Accuracy_sparsemax_tencent.pdf_tex
\begingroup%
  \makeatletter%
  \providecommand\color[2][]{%
    \errmessage{(Inkscape) Color is used for the text in Inkscape, but the package 'color.sty' is not loaded}%
    \renewcommand\color[2][]{}%
  }%
  \providecommand\transparent[1]{%
    \errmessage{(Inkscape) Transparency is used (non-zero) for the text in Inkscape, but the package 'transparent.sty' is not loaded}%
    \renewcommand\transparent[1]{}%
  }%
  \providecommand\rotatebox[2]{#2}%
  \ifx\svgwidth\undefined%
    \setlength{\unitlength}{192.08655954bp}%
    \ifx\svgscale\undefined%
      \relax%
    \else%
      \setlength{\unitlength}{\unitlength * \real{\svgscale}}%
    \fi%
  \else%
    \setlength{\unitlength}{\svgwidth}%
  \fi%
  \global\let\svgwidth\undefined%
  \global\let\svgscale\undefined%
  \makeatother%
  \begin{picture}(1,0.4763636)%
    \put(0,0){\includegraphics[width=\unitlength,page=1]{barplot_Accuracy_sparsemax_tencent.pdf}}%
  \end{picture}%
\endgroup%

%% file: charts/barplot_MacroF1_sparsemax_tencent.pdf_tex
\begingroup%
  \makeatletter%
  \providecommand\color[2][]{%
    \errmessage{(Inkscape) Color is used for the text in Inkscape, but the package 'color.sty' is not loaded}%
    \renewcommand\color[2][]{}%
  }%
  \providecommand\transparent[1]{%
    \errmessage{(Inkscape) Transparency is used (non-zero) for the text in Inkscape, but the package 'transparent.sty' is not loaded}%
    \renewcommand\transparent[1]{}%
  }%
  \providecommand\rotatebox[2]{#2}%
  \ifx\svgwidth\undefined%
    \setlength{\unitlength}{191.98155954bp}%
    \ifx\svgscale\undefined%
      \relax%
    \else%
      \setlength{\unitlength}{\unitlength * \real{\svgscale}}%
    \fi%
  \else%
    \setlength{\unitlength}{\svgwidth}%
  \fi%
  \global\let\svgwidth\undefined%
  \global\let\svgscale\undefined%
  \makeatother%
  \begin{picture}(1,0.47662414)%
    \put(0,0){\includegraphics[width=\unitlength,page=1]{barplot_MacroF1_sparsemax_tencent.pdf}}%
  \end{picture}%
\endgroup%